%% file: main.tex
\pdfoutput=1

\documentclass[11pt]{article}

\usepackage[]{naacl2021}

\usepackage{times}
\usepackage{latexsym}

\usepackage{graphicx}
\usepackage{subcaption}
\usepackage{multirow}
\usepackage{amsmath}
\usepackage{diagbox}
\usepackage[ruled,vlined]{algorithm2e}
\usepackage{pgfplots}

\usepackage{url}
\usepackage{xcolor}
\usepackage{todonotes} 

\pgfplotsset{compat=1.17}

\usepackage[T1]{fontenc}

\usepackage[utf8]{inputenc}

\usepackage{microtype}

\title{Learning Policies for Multilingual Training of Neural Machine Translation Systems}

\author{Gaurav Kumar, Philipp Koehn, Sanjeev Khudanpur \\
  Center for Language and Speech Processing \\
  Johns Hopkins University \\
  {\tt gkumar@cs.jhu.edu},
  {\tt \{phi, khudanpur\}}\tt{@jhu.edu} \\
  }

\begin{document}
\maketitle
\begin{abstract}
Low-resource Multilingual Neural Machine Translation (MNMT) is typically tasked with improving the translation performance on one or more language pairs with the aid of high-resource language pairs. In this paper, we propose two simple search based curricula -- orderings of the multilingual training data -- which help improve translation performance in conjunction with existing techniques such as fine-tuning. Additionally, we attempt to learn a curriculum for MNMT from scratch jointly with the training of the translation system with the aid of contextual multi-arm bandits. We show on the FLORES low-resource translation dataset that these learned curricula can provide better starting points for fine tuning and improve overall performance of the translation system. 
\end{abstract}

\section{Introduction}
Curriculum learning \cite{BengioLouradourCollobertWeston2009, ELMAN199371, Rhode99} hypothesizes that presenting training samples in a meaningful order to machine learners during training may help improve model quality and convergence speed.
In the field of Neural Machine Translation (NMT) most curricula are hand designed e.g., fine-tuning \cite{Luong-Manning:iwslt15, DBLP:journals/corr/FreitagA16} and data selection \cite{Moore:2010, Axelrod:2011:DAV:2145432.2145474, duh13acl, DBLP:conf/coling/DurraniSJA16}. Another common curriculum is one based on ordering samples from \emph{easy} to \emph{hard} using linguistic features and auxiliary model scores \cite{Zhang18, Zhang2019NAACL} but these are hard to tune, relying to extensive trial and error to find the right hyperparameters. Attempts to learn a curriculum jointly with the NMT training setup \cite{kumar-etal-2019-reinforcement} can suffer from observation sparsity, where a single training run does not provide enough training samples for an external agent to learn a good curriculum policy.

Our NMT task of choice in this paper will be \emph{low-resource multi-lingual NMT} (MNMT). While standard NMT systems typically deal with a language pair, the source and the target, an MNMT model may have multiple languages as source and/or target. Most large-scale MNMT models are trained using some form of model parameter sharing \cite{johnson-etal-2017-googles, aharoni-etal-2019-massively, DBLP:journals/corr/abs-1907-05019, bapna-firat-2019-simple}. The notion of how input data should be presented to the MNMT system during training only finds prominence in the case of low-resource MNMT. A typical low-resource task will try to leverage a high-resource language pair to aid the training of an NMT system for a low-resource (very small or no parallel data available) and related language-pair of interest. Typical approaches for low resource MNMT involve pivoting and zero-shot training \cite{DBLP:journals/corr/abs-1811-01389, johnson-etal-2017-googles} and transfer learning via fine-tuning \cite{zoph-etal-2016-transfer, dabre-etal-2019-exploiting}. \newcite{DBLP:journals/corr/FinnAL17} attempt to meta-learn parameter initialization for child models using trained-high resource parent models for this task.

In this paper, we build upon the framework for learning curricula and attempt to alleviate the problem of observation sparsity by learning more robust policies from multiple training runs. We use contextual multi-arm bandits for our agents which learn multilingual data sampling policies jointly with the training of the NMT system. Additionally, we explore some simple policy search methods to our list of baselines; specifically, we try and find the best policies using the expensive grid search and pruned-tree search methods. We use state-of-the-art hand-designed curricula as our baselines to beat. Building upon the task and datasets established by \newcite{guzman-etal-2019-flores}, in this paper, we will attempt to learn a curriculum to train an NMT system for the Nepali-English language pair while leveraging the high resource Hindi-English pair. The agent will learn to choose between mini-batches containing either Hindi-English or Nepali-English data at each time step during NMT training to maximize the expected reward (improvement in validation set performance). The learned curriculum will hence condition on the state of the NMT system during training and determine whether to expose it to a batch of Nepali-English or Hindi-English data. We start by presenting our methods for obtaining search-based and learned curricula in section~\ref{sec:methods}. We present our experiment setup in section~\ref{sec:experiment} and results in section~\ref{sec:results}.

\section{Methods}
\label{sec:methods}
\begin{figure}[t]
    \centering
    \includegraphics[width=\linewidth]{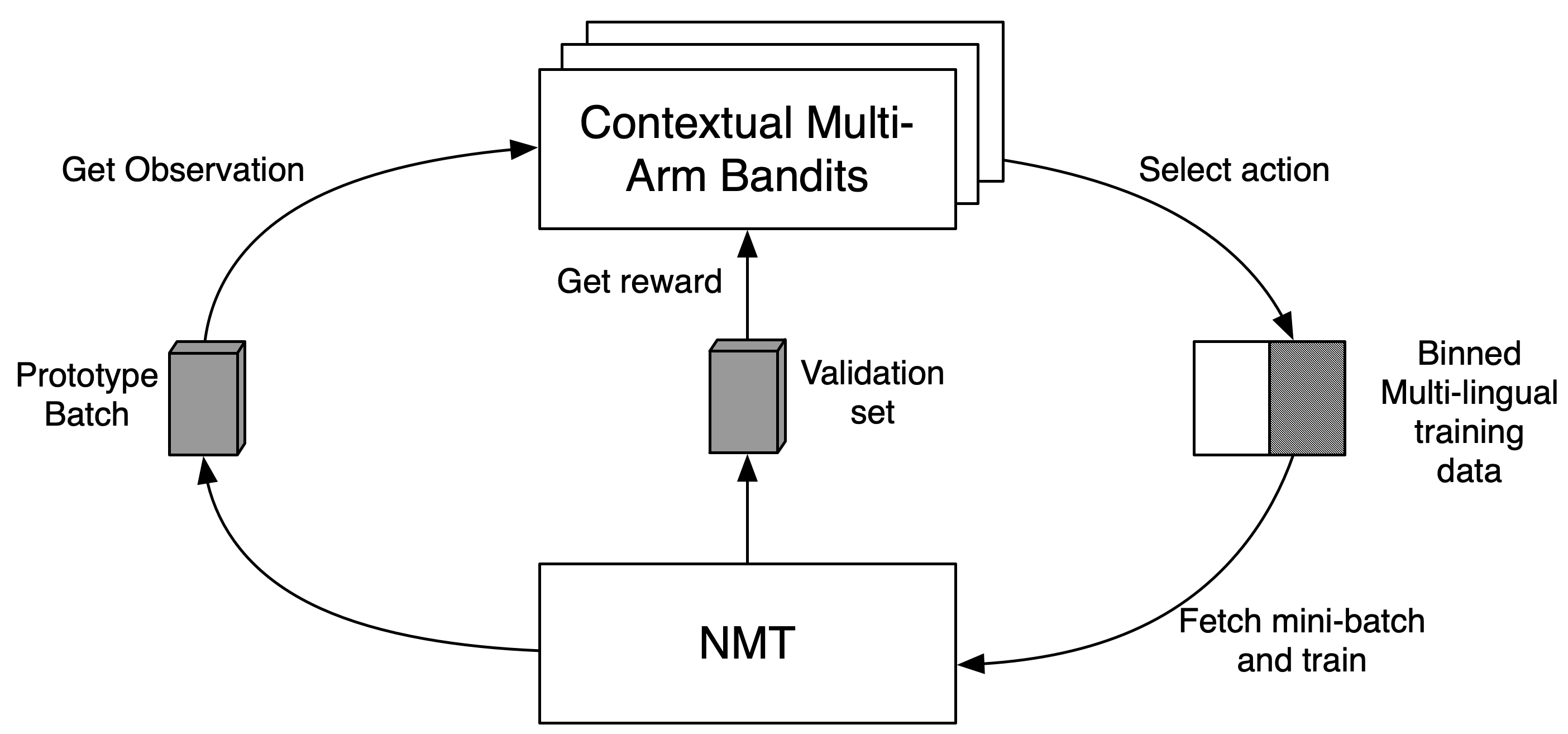}
    \caption{The multi-arm bandit agents' (MAB) interface with the NMT system.}
    \label{fig:bandit-interface}
\end{figure}

The procedure for learning a two-bin policy for multi-lingual training uses multiple multi-arm bandits which explore independent of each other and effectively learn their own policies. The stochastic nature of the exploration policy ensures that they explore different spaces to the observation-reward space. Figure~\ref{fig:bandit-interface} shows an overview of this interface. The training data for all agents is pooled at the end of the training of individual agents and one final agent is trained using this data which determines the final policy we use as our multi-lingual curriculum.

\subsection{Data Binning}
Instead of mixing together all the language pairs into one single dataset, we create separate batches for each language pair. Hence, with respect to the agent, this is a two bin problem, where its \emph{action} is the choice of the bin to draw a mini-batch. As a result of this design decision, each batch will only contain a single language pair. More generally, this can be extended to an arbitrary number of bins, one per language-pair being used to train the MNMT system.

\subsection{Observation Engineering}
The observations provided to the multi-arm bandits are identical to the ones introduced \newcite{kumar-etal-2019-reinforcement}. A \emph{prototype} batch -- a finite number of sentences from each language pair -- is sampled per bin (language-pair) and concatenated together. At each time step, the observation is the vector containing sentence-level log-likelihoods produced by the NMT system for this prototype batch. We exclude observations from the initial portion of NMT interaction to counteract the naturally decaying property of log-likelihood scores during NMT training. 

\subsection{Grid-search baselines}
The simplest (albeit expensive to find) search-based learn-able curriculum to consider in this case is one where we sample batches from one language with a fixed probability or else sample from the other bin during training. Since there is only one degree of freedom in this search problem, we perform a simple line-search over the range of possible values for this probability. Note that, although this curriculum is `learned' it remains fixed during training and does not change based on the state of the NMT system.

\begin{algorithm}[tbp]
\SetAlgoLined
\KwResult{$p^*$, the list of the best policies per phase}
 $\hat{p} = \{0.0, 0.1, \cdots, 1.0\}$ // Policies to explore\;
 Randomly initialize starting NMT model $\Theta^*$\;
 \While{NMT next training phase $t$ exists}{
  \For{$p$ in $p^*$} {
    Bin sampling probablity = $p$\;
    Training start checkpoint = $\Theta^*$\;
    Run training of NMT training for phase $t$;\;
    Store trained model checkpoint $\theta$
  } 
  Select model $\theta^*$ with best score on validation set\;
  $p^* = p^* + [(t, p)]$\;
  $\Theta^* = \theta^*$\;
 }
 \caption{Pruned tree-search for multi-lingual curricula search}
 \label{alg:pruned_search}
\end{algorithm}

\subsection{Pruned Tree search}
A variation of the previous search method involves one which uses a technique similar to beam search. We divide training into a finite number of \emph{phases} and then starting from the beginning of training, we search for the best fixed sampling probability. At the end of this \emph{phase}, we discard all but the best model and the policy which led to it, and continue the search for the best policy in the next phase from this model checkpoint. The result is a tree-search which prunes all but the best node after each phase. The final policy is the culmination of all phase-wise best fixed sampling ratios. This procedure appears in Algorithm~\ref{alg:pruned_search}.

\subsection{Contextual Multi-arm Bandits}
Multi-arm bandit (MAB) based agents are typically trained to learn policies which maximize the expected reward received (minimize regret). Contextual multi-arm bandits \cite{Pandey2007, 1406128, NIPS2007_4b04a686} allows the use of state based information to determine this policy. In our case the contextual MABs condition on the \emph{observation} received from the NMT system to determine an \emph{action}, the choice of bin to sample a mini-batch. The \emph{reward} obtained for this action is the \emph{delta-validation perplexity} post update as described in \newcite{kumar-etal-2019-reinforcement}. The exploration strategy is the linearly-decaying \emph{epsilon-greedy} strategy \cite{DBLP:journals/corr/KuleshovP14}. The contextual MABs are implemented as simple feed-forward neural networks which take the \emph{observation} vector as input and produce a distribution over two states representing the bins. If we choose to exploit this learned policy, the bin with maximum probability mass is selected for sampling.

\section{Experiment Setup}
\label{sec:experiment}
We use Fairseq \cite{ott2019fairseq} for all our NMT experiments and the our NMT systems are configured to replicate the setup described in \newcite{guzman-etal-2019-flores}. The grid search experiments search over the the range $[0, 1]$ for sampling in increments of 0.1. The pruned tree-search uses a beam width of 1. The phase duration for tree-search is set to one epoch of NMT training. We use either 5 or 10 concurrent contextual MABs which are implemented as two 256-dimensional feed forward neural networks trained using RMSProp with a learning rate of 0.00025 and a decay of 0.95 and no momentum. Rewards for the agent (validation delta-perplexity) are provided every ten training steps. item Observations: We sample 32 prototype sentences from each bin to create a {\em prototype batch} of 64 sentences. We use an NMT warmup of 5000 steps (no transitions from this period are recorded). For the exploration strategy we use a linearly decaying epsilon function with decay period set to 25k steps. The decay floor was set to 0.01.
The window for the delta-perplexity reward was 1.

\begin{table}[tbp]
\centering
\begin{tabular}{|l|l|l|}
\hline
Dataset        & Sentences & Tokens \\ \hline
Nepali-English & 563K      & 6.8M   \\ \hline
Hindi-English  & 1.6M      & 16.7M  \\ \hline
\end{tabular}
\caption{Statistics of the training data for the Nepali-Hindi-English multilingual NMT system.}
\label{tab:c5_data_stats}
\end{table}

We use the datasets provided as part of the FLORES task \cite{guzman-etal-2019-flores} for our experiments. The statistics of the training dataset for the multi-lingual task appear in table~\ref{tab:c5_data_stats}. The Hindi-English dataset comes from the IIT Bombay corpus\footnote{\url{ http://www.cfilt.iitb.ac.in/iitb_parallel}}. The validation and test sets for Nepali-English (the low resource language-pair of interest) contain 2500 and 3000 sentences respectively.

\section{Results}
\label{sec:results}
\begin{table*}[tbp]
\centering
\begin{tabular}{|l|l|l|}
\hline
                                                & valid          & test           \\ \hline
\multicolumn{3}{|c|}{Baselines}                                                   \\ \hline
ne-en: Random Baseline                          & 6.35           & 7.71           \\ \hline
hi-en: Random baseline (with ne valid)          & 2.71           & 3.9            \\ \hline
ne-hi-en: Random Baseline                       & \textbf{12.24} & \textbf{14.88} \\ \hline
ne-hi-en: Multi-lingual Transformer             & 12.01          & 14.78          \\ \hline
ne-hi-en: Continued training from hi-en         & 12.2           & 14.3           \\ \hline
\multicolumn{3}{|c|}{Searched Curricula}                                           \\ \hline
Grid Search (best = 50/50)                      & 12.01          & 14.78          \\ \hline
Grid Search (best = 50/50) + Continued training & \textbf{12.33} & \textbf{15.1}  \\ \hline
Pruned Tree-search                              & 12.3           & 14.8           \\ \hline
Pruned Tree-search + Continued training         & 12.41          & 14.92          \\ \hline
\multicolumn{3}{|c|}{Agent Learned Curricula}                                      \\ \hline
MAB1 (best = 10 concurrent, 500 updates)        & \textbf{12.21} & \textbf{14.87} \\ \hline
MAB2 (best = 5 concurrent, 2 epochs)            & 12.18          & 14.67          \\ \hline
\end{tabular}
\caption{BLEU scores for the Nepali-English test set using the fixed, searched and learned multilingual curricula.}
\label{tab:c5_results}
\end{table*}

Our results are presented in Table~\ref{tab:c5_results}. Our baselines consist of:
\begin{itemize}
    \item ne-en random baseline: This is the NMT setup which is only trained on the Nepali-English corpus. The data is randomly shuffled to form mini-batches.
    \item hi-en random baseline: The NMT system trained on the high-resource Hindi-English dataset with the Nepali-English validation and test sets.
    \item ne-hi-en random baseline: The Hindi-English and Nepali-English data is mixed together to train the NMT system. The Nepali-English data is upsampled to match the size of the the Hindi-English corpus.
    \item Multilingual transformer: Replicates the setup from \newcite{guzman-etal-2019-flores}.
    \item Continued training baseline: Uses the hi-en random baseline as a starting point to fine tune using the Nepali-English validation and test sets.
\end{itemize}

Our non-MAB search-based curriculum baselines are:
\begin{itemize}
    \item Grid search: A static curriculum is learned by searching over the space of sampling probabilities for the bins.
    \item Grid Search + Continued training: The previous model is fine tuned using the Nepali-English validation and test sets.
    \item Pruned tree-search: Epoch-dependent curriculum searched using the pruned tree-search method.
    \item Pruned tree-search + Continued training: The previous model is fine tuned using the Nepali-English validation and test sets.
\end{itemize}

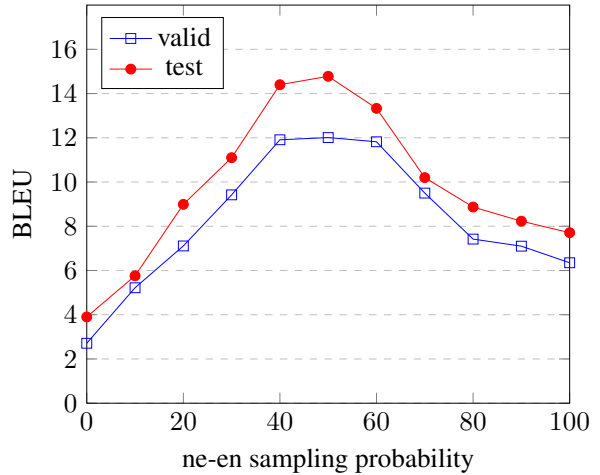
\begin{figure}[tbp]
\centering
\input{img/grid_search_fig}    
\caption{BLEU scores for the Nepali-English validation and test set at various values of the ne-en sampling probability.}
\label{fig:grid_search}
\end{figure}

From Table~\ref{tab:c5_results}, we see that the ne-en and hi-en baselines are very weak, with the latter lagging behind despite having access to more data. This indicates that with these language pairs, even though adding the high-resource dataset may help, in isolation it is not a good proxy for the low-resource pair. The random baseline with the combination of the two datasets (upsampled low-resource) is the strongest amongst the fixed baselines marginally beating the multi-lingual transformer and the (surprisingly) the continued training baselines. While the grid search and pruned-tree search baselines are close in performance to the best fixed baselines, continued training with them provides much stronger results where the 50/50 configuration for the grid search provides the best result at 15.1 BLEU and the tree search slightly behind at 14.92 BLEU. Figure~\ref{fig:grid_search} shows the BLEU scores for the grid search experiments over the chosen search points in the probability space.

\begin{table}[tbp]
\centering
\begin{tabular}{|l|l|l|}
\hline
                                                        & valid          & test           \\ \hline
MAB (5 conc., 500 updates)                         & 12.2           & 14.11          \\ \hline
MAB (10 conc., 500 updates)                        & \textbf{12.21} & \textbf{14.87} \\ \hline
MAB (5 conc., 1 epoch)                             & 11.44          & 13.98          \\ \hline
MAB (5 conc., 2 epoch)                             & 12.18          & 14.67          \\ \hline
\end{tabular}
\caption{BLEU scores for the Nepali-English test set using various configurations of the contextual MABs to learn the multilingual sampling curriculum.}
\label{tab:mab_results}
\end{table}

For the contextual MABs, we use either 5 or 10 concurrent agents (training data is gathered from all concurrent bandits to train the final curriculum). In addition, we choose to update the bandit policy only once every 500 updates, 1 epoch or 2 epochs of NMT training. The results of all our experiments appear in table~\ref{tab:mab_results} and the best configurations are in table~\ref{tab:c5_results}. While the curricula learned using the contextual MABs are able to match the performance of the strongest fixed policy (ne-hi-en random baseline), it performs slightly worse than the curriculum obtained using the (expensive) grid search combined with continued training, by about 0.2 BLEU points. 

\section{Conclusion}
In this paper, we build upon the approach we present techinques which learn curricula for multilingual NMT training from multiple training runs. On the task of low-resource multilingual NMT training, we learn a curriculum using conditional multi-arm bandits which conditions on the state of the NMT system and decides to either train on a batch of a high-resource (Hindi-English) or the low-resource (Nepali-Hindi) language pair. In addition, we introduce some simple search-based methods for policy search (grid search and pruned tree search) for this task. We show that both these simple \emph{learned} curricula and the ones derived from the MABs can match the state-of-the-art hand-designed multilingual baselines. However, continued training with these \emph{learned} curricula provide slightly better results, indicating that they may serve as good starting models for fine-tuning (another possible benefit of curriculum learning).

\clearpage

\bibliography{anthology,custom}
\bibliographystyle{acl_natbib}

\end{document}

%% file: img/grid_search_fig.tex
\resizebox {0.5\textwidth} {!} {
\begin{tikzpicture}
\begin{axis}[
    xlabel={ne-en sampling probability},
    ylabel={BLEU},
    xmin=0, xmax=100,
    ymin=0, ymax=18,
    xtick={0,20,40,60,80,100},
    ytick={0,2,4,6,8,10,12,14,16},
    legend pos=north west,
    ymajorgrids=true,
    grid style=dashed,
]

\addplot[
    color=blue,
    mark=square,
    ]
    coordinates {
    (0,2.71)(10,5.22)(20,7.11)(30,9.42)(40,11.91)(50,12.01)(60,11.82)(70,9.5)(80,7.42)(90,7.1)(100,6.35)
    };
    
\addplot[
    color=red,
    mark=*,
    ]
    coordinates {
    (0,3.9)(10,5.76)(20,8.99)(30,11.1)(40,14.4)(50,14.78)(60,13.33)(70,10.2)(80,8.87)(90,8.23)(100,7.71)
    };
    \addlegendentry{valid}
    \addlegendentry{test}
    
\end{axis}
\end{tikzpicture}
}

%% file: main.bbl
\begin{thebibliography}{26}
\expandafter\ifx\csname natexlab\endcsname\relax\def\natexlab#1{#1}\fi

\bibitem[{Aharoni et~al.(2019)Aharoni, Johnson, and
  Firat}]{aharoni-etal-2019-massively}
Roee Aharoni, Melvin Johnson, and Orhan Firat. 2019.
\newblock \href {https://doi.org/10.18653/v1/N19-1388} {Massively multilingual
  neural machine translation}.
\newblock In \emph{Proceedings of the 2019 Conference of the North {A}merican
  Chapter of the Association for Computational Linguistics: Human Language
  Technologies, Volume 1 (Long and Short Papers)}, pages 3874--3884,
  Minneapolis, Minnesota. Association for Computational Linguistics.

\bibitem[{Arivazhagan et~al.(2019)Arivazhagan, Bapna, Firat, Lepikhin, Johnson,
  Krikun, Chen, Cao, Foster, Cherry, Macherey, Chen, and
  Wu}]{DBLP:journals/corr/abs-1907-05019}
Naveen Arivazhagan, Ankur Bapna, Orhan Firat, Dmitry Lepikhin, Melvin Johnson,
  Maxim Krikun, Mia~Xu Chen, Yuan Cao, George~F. Foster, Colin Cherry, Wolfgang
  Macherey, Zhifeng Chen, and Yonghui Wu. 2019.
\newblock \href {http://arxiv.org/abs/1907.05019} {Massively multilingual
  neural machine translation in the wild: Findings and challenges}.
\newblock \emph{CoRR}, abs/1907.05019.

\bibitem[{Axelrod et~al.(2011)Axelrod, He, and
  Gao}]{Axelrod:2011:DAV:2145432.2145474}
Amittai Axelrod, Xiaodong He, and Jianfeng Gao. 2011.
\newblock \href {http://dl.acm.org/citation.cfm?id=2145432.2145474} {Domain
  adaptation via pseudo in-domain data selection}.
\newblock In \emph{Proceedings of the Conference on Empirical Methods in
  Natural Language Processing}, EMNLP '11, pages 355--362. Association for
  Computational Linguistics.

\bibitem[{Bapna and Firat(2019)}]{bapna-firat-2019-simple}
Ankur Bapna and Orhan Firat. 2019.
\newblock \href {https://doi.org/10.18653/v1/D19-1165} {Simple, scalable
  adaptation for neural machine translation}.
\newblock In \emph{Proceedings of the 2019 Conference on Empirical Methods in
  Natural Language Processing and the 9th International Joint Conference on
  Natural Language Processing (EMNLP-IJCNLP)}, pages 1538--1548, Hong Kong,
  China. Association for Computational Linguistics.

\bibitem[{Bengio et~al.(2009)Bengio, Louradour, Collobert, and
  Weston}]{BengioLouradourCollobertWeston2009}
Yoshua Bengio, J\'er\^ome Louradour, Ronan Collobert, and Jason Weston. 2009.
\newblock \href {https://doi.org/10.1145/1553374.1553380} {Curriculum
  {{Learning}}}.
\newblock In \emph{Proceedings of the 26th {{Annual International Conference}}
  on {{Machine Learning}}}, ICML '09, pages 41--48, Montreal, Quebec, Canada.
  {ACM}.

\bibitem[{{Chih-Chun Wang} et~al.(2005){Chih-Chun Wang}, {Kulkarni}, and
  {Poor}}]{1406128}
{Chih-Chun Wang}, S.~R. {Kulkarni}, and H.~V. {Poor}. 2005.
\newblock \href {https://doi.org/10.1109/TAC.2005.844079} {Bandit problems with
  side observations}.
\newblock \emph{IEEE Transactions on Automatic Control}, 50(3):338--355.

\bibitem[{Dabre et~al.(2019)Dabre, Fujita, and
  Chu}]{dabre-etal-2019-exploiting}
Raj Dabre, Atsushi Fujita, and Chenhui Chu. 2019.
\newblock \href {https://doi.org/10.18653/v1/D19-1146} {Exploiting
  multilingualism through multistage fine-tuning for low-resource neural
  machine translation}.
\newblock In \emph{Proceedings of the 2019 Conference on Empirical Methods in
  Natural Language Processing and the 9th International Joint Conference on
  Natural Language Processing (EMNLP-IJCNLP)}, pages 1410--1416, Hong Kong,
  China. Association for Computational Linguistics.

\bibitem[{Duh et~al.(2013)Duh, Neubig, Sudoh, and Tsukada}]{duh13acl}
Kevin Duh, Graham Neubig, Katsuhito Sudoh, and Hajime Tsukada. 2013.
\newblock \href {http://www.phontron.com/paper/duh13acl.pdf} {Adaptation data
  selection using neural language models: Experiments in machine translation}.
\newblock In \emph{The 51st Annual Meeting of the Association for Computational
  Linguistics (ACL)}, pages 678--683, Sofia, Bulgaria.

\bibitem[{Durrani et~al.(2016)Durrani, Sajjad, Joty, and
  Abdelali}]{DBLP:conf/coling/DurraniSJA16}
Nadir Durrani, Hassan Sajjad, Shafiq~R. Joty, and Ahmed Abdelali. 2016.
\newblock A deep fusion model for domain adaptation in phrase-based {MT}.
\newblock In \emph{{COLING}}, pages 3177--3187. {ACL}.

\bibitem[{Elman(1993)}]{ELMAN199371}
Jeffrey~L. Elman. 1993.
\newblock \href {https://doi.org/https://doi.org/10.1016/0010-0277(93)90058-4}
  {Learning and development in neural networks: the importance of starting
  small}.
\newblock \emph{Cognition}, 48(1):71 -- 99.

\bibitem[{Finn et~al.(2017)Finn, Abbeel, and
  Levine}]{DBLP:journals/corr/FinnAL17}
Chelsea Finn, Pieter Abbeel, and Sergey Levine. 2017.
\newblock \href {http://arxiv.org/abs/1703.03400} {Model-agnostic meta-learning
  for fast adaptation of deep networks}.
\newblock \emph{CoRR}, abs/1703.03400.

\bibitem[{Freitag and Al{-}Onaizan(2016)}]{DBLP:journals/corr/FreitagA16}
Markus Freitag and Yaser Al{-}Onaizan. 2016.
\newblock \href {http://arxiv.org/abs/1612.06897} {Fast domain adaptation for
  neural machine translation}.
\newblock \emph{CoRR}, abs/1612.06897.

\bibitem[{Guzm{\'a}n et~al.(2019)Guzm{\'a}n, Chen, Ott, Pino, Lample, Koehn,
  Chaudhary, and Ranzato}]{guzman-etal-2019-flores}
Francisco Guzm{\'a}n, Peng-Jen Chen, Myle Ott, Juan Pino, Guillaume Lample,
  Philipp Koehn, Vishrav Chaudhary, and Marc{'}Aurelio Ranzato. 2019.
\newblock \href {https://doi.org/10.18653/v1/D19-1632} {The {FLORES} evaluation
  datasets for low-resource machine translation: {N}epali{--}{E}nglish and
  {S}inhala{--}{E}nglish}.
\newblock In \emph{Proceedings of the 2019 Conference on Empirical Methods in
  Natural Language Processing and the 9th International Joint Conference on
  Natural Language Processing (EMNLP-IJCNLP)}, pages 6098--6111, Hong Kong,
  China. Association for Computational Linguistics.

\bibitem[{Johnson et~al.(2017)Johnson, Schuster, Le, Krikun, Wu, Chen, Thorat,
  Viegas, Wattenberg, Corrado, Hughes, and Dean}]{johnson-etal-2017-googles}
Melvin Johnson, Mike Schuster, Quoc~V. Le, Maxim Krikun, Yonghui Wu, Zhifeng
  Chen, Nikhil Thorat, Fernanda Viegas, Martin Wattenberg, Greg Corrado,
  Macduff Hughes, and Jeffrey Dean. 2017.
\newblock \href {https://doi.org/10.1162/tacl_a_00065} {{G}oogle{'}s
  multilingual neural machine translation system: Enabling zero-shot
  translation}.
\newblock \emph{Transactions of the Association for Computational Linguistics},
  5:339--351.

\bibitem[{Kuleshov and Precup(2014)}]{DBLP:journals/corr/KuleshovP14}
Volodymyr Kuleshov and Doina Precup. 2014.
\newblock \href {http://arxiv.org/abs/1402.6028} {Algorithms for multi-armed
  bandit problems}.
\newblock \emph{CoRR}, abs/1402.6028.

\bibitem[{Kumar et~al.(2019)Kumar, Foster, Cherry, and
  Krikun}]{kumar-etal-2019-reinforcement}
Gaurav Kumar, George Foster, Colin Cherry, and Maxim Krikun. 2019.
\newblock \href {https://doi.org/10.18653/v1/N19-1208} {Reinforcement learning
  based curriculum optimization for neural machine translation}.
\newblock In \emph{Proceedings of the 2019 Conference of the North {A}merican
  Chapter of the Association for Computational Linguistics: Human Language
  Technologies, Volume 1 (Long and Short Papers)}, pages 2054--2061,
  Minneapolis, Minnesota. Association for Computational Linguistics.

\bibitem[{Lakew et~al.(2018)Lakew, Lotito, Negri, Turchi, and
  Federico}]{DBLP:journals/corr/abs-1811-01389}
Surafel~Melaku Lakew, Quintino~F. Lotito, Matteo Negri, Marco Turchi, and
  Marcello Federico. 2018.
\newblock \href {http://arxiv.org/abs/1811.01389} {Improving zero-shot
  translation of low-resource languages}.
\newblock \emph{CoRR}, abs/1811.01389.

\bibitem[{Langford and Zhang(2008)}]{NIPS2007_4b04a686}
John Langford and Tong Zhang. 2008.
\newblock \href
  {https://proceedings.neurips.cc/paper/2007/file/4b04a686b0ad13dce35fa99fa4161c65-Paper.pdf}
  {The epoch-greedy algorithm for multi-armed bandits with side information}.
\newblock In \emph{Advances in Neural Information Processing Systems},
  volume~20. Curran Associates, Inc.

\bibitem[{Luong and Manning(2015)}]{Luong-Manning:iwslt15}
Minh-Thang Luong and Christopher~D. Manning. 2015.
\newblock Stanford neural machine translation systems for spoken language
  domain.
\newblock In \emph{International Workshop on Spoken Language Translation}, Da
  Nang, Vietnam.

\bibitem[{Moore and Lewis(2010)}]{Moore:2010}
Robert~C. Moore and William Lewis. 2010.
\newblock \href {http://dl.acm.org/citation.cfm?id=1858842.1858883}
  {Intelligent selection of language model training data}.
\newblock In \emph{Proceedings of the ACL 2010 Conference Short Papers},
  ACLShort '10, pages 220--224. Association for Computational Linguistics.

\bibitem[{Ott et~al.(2019)Ott, Edunov, Baevski, Fan, Gross, Ng, Grangier, and
  Auli}]{ott2019fairseq}
Myle Ott, Sergey Edunov, Alexei Baevski, Angela Fan, Sam Gross, Nathan Ng,
  David Grangier, and Michael Auli. 2019.
\newblock fairseq: A fast, extensible toolkit for sequence modeling.
\newblock In \emph{Proceedings of NAACL-HLT 2019: Demonstrations}.

\bibitem[{Pandey et~al.(2007)Pandey, Agarwal, Chakrabarti, and
  Josifovski}]{Pandey2007}
S.~Pandey, D.~Agarwal, D.~Chakrabarti, and V.~Josifovski. 2007.
\newblock Bandits for taxonomies: a model-based approach.
\newblock In \emph{SIAM Data Mining Conference}.

\bibitem[{Rohde and Plaut(1994)}]{Rhode99}
D.~L. Rohde and D.~C. Plaut. 1994.
\newblock Language acquisition in the absence of explicit negative evidence:
  how important is starting small?
\newblock In \emph{Cognition}, volume~72, pages 67--109.

\bibitem[{Zhang et~al.(2018)Zhang, Kumar, Khayrallah, Murray, Gwinnup,
  Martindale, McNamee, Duh, and Carpuat}]{Zhang18}
Xuan Zhang, Gaurav Kumar, Huda Khayrallah, Kenton Murray, Jeremy Gwinnup,
  Marianna~J. Martindale, Paul McNamee, Kevin Duh, and Marine Carpuat. 2018.
\newblock \href {http://arxiv.org/abs/1811.00739} {An empirical exploration of
  curriculum learning for neural machine translation}.
\newblock \emph{CoRR}, abs/1811.00739.

\bibitem[{Zhang et~al.(2019)Zhang, Shapiro, Kumar, McNamee, Carpuat, and
  Duh}]{Zhang2019NAACL}
Xuan Zhang, Pamela Shapiro, Gaurav Kumar, Paul McNamee, Marine Carpuat, and
  Kevin Duh. 2019.
\newblock Curriculum learning for domain adaptation in neural machine
  translation.
\newblock In \emph{Proceedings of the 2019 Conference of the North American
  Chapter of the Association for Computational Linguistics: Human Language
  Technologies, Volume 1 (Long Papers)}. Association for Computational
  Linguistics.

\bibitem[{Zoph et~al.(2016)Zoph, Yuret, May, and
  Knight}]{zoph-etal-2016-transfer}
Barret Zoph, Deniz Yuret, Jonathan May, and Kevin Knight. 2016.
\newblock \href {https://doi.org/10.18653/v1/D16-1163} {Transfer learning for
  low-resource neural machine translation}.
\newblock In \emph{Proceedings of the 2016 Conference on Empirical Methods in
  Natural Language Processing}, pages 1568--1575, Austin, Texas. Association
  for Computational Linguistics.

\end{thebibliography}
